# Hybrid Fuzzy-Linear Programming Approach for Multi Criteria Decision Making Problems

Sonja Petrovic-Lazarevic[1] and Ajith Abraham[2]
[1]Monash University, Department of Management, Australia
[2]Department of Computer Science, Oklahoma State University (Tulsa), Oklahoma 74106, USA, Email: ajith.abraham@ieee.org

**Abstract**
The purpose of this paper is to point to the usefulness of applying a linear mathematical formulation of fuzzy multiple criteria objective decision methods in organising business activities. In this respect fuzzy parameters of linear programming are modelled by preference-based membership functions. This paper begins with an introduction and some related research followed by some fundamentals of fuzzy set theory and technical concepts of fuzzy multiple objective decision models. Further a real case study of a manufacturing plant and the implementation of the proposed technique is presented. Empirical results clearly show the superiority of the fuzzy technique in optimising individual objective functions when compared to non-fuzzy approach. Furthermore, for the problem considered, the optimal solution helps to infer that by incorporating fuzziness in a linear programming model either in constraints, or both in objective functions and constraints, provides a similar (or even better) level of satisfaction for obtained results compared to non-fuzzy linear programming.

**Keywords:** Fuzzy multiple objective decision making, business activities, degree of satisfaction, tolerances of fuzzy constraints

## 1. INTRODUCTION

Many years before the introduction of mathematical planning methods decision processes of organising business activities were based on intuition and experience. Decisions were subject to professional judgements usually based on imprecise information. Today business-planning processes in spite of the application of mathematical planning principles still utilise subjective judgements, making decisions often vague. Organising a business activity is a multiple objective decision process. Since a decision is usually vague, it may be based on fuzzy numbers. In 1980 Dyson stated that fuzzy programming models should not be treated as a new contribution to multiple objective decision making methods, but rather as a lead to new conventional decision methods. "Support for this thesis would require examples of new and affective fuzzy inspired multi-criteria methods" (Dyson, 1980). The paper is an attempt to point to a significance of applying a fuzzy approach to multi objective decision methods in the process of organising business activities. In this respect the paper analyses the appropriateness of applying either the non-fuzzy multi objective decision model, with crisp objectives and fuzzy constraints, or the fuzzy multi objective decision model, with both fuzzy objectives and fuzzy constraints. The paper is organised as follows: next section covers a literature review of multi objective decision models. Section three contains definitions relevant to fuzzy set theory. Section four elaborates non-fuzzy multiple objective decision model (MODM) and the fuzzy MODM (FMODM) is explained in section five. The results of two models are presented in the case study of a construction firm that produces, transports and places concrete on construction sites. The paper ends with concluding remarks and future research directions.

## 2. LITERATURE REVIEW

In real-world decision-making processes in business, decision making theory has become one of most important fields. It uses the optimisation methodology connected with a single criterion, but also satisfying concepts of multiple criteria. Decision processes with multiple criteria deal with human judgement. This is not easy to model. The human judgement element is in the area of preferences defined by the decision maker (Chankong & Haimes, 1983) First attempts to model decision processes with multiple criteria in business lead to concepts of goal programming (Ignizio, 1976). In this approach the decision maker underpins each objective with a number of goals that should be satisfied (Lootsma, 1989) Satisfying requires finding a solution to a multi criterion problem, which is preferred, understood and implemented with confidence. The confidence that the best solution has been found is estimated through the " ideal solution". That is the solution which optimises all criteria simultaneously. Since this is practically unattainable a decision maker considers feasible solutions closest to the ideal solution (Zeleny, 1982).

In goal programming the preferences required from the decision maker are presented with weights, targets, trade-offs and goal levels to formulate the problem. Steuer (1986) proposed the objective function of a linear goal programming to be a weighting representation of second objective functions with the sum of these weights equal to unity. Allowing these weights to vary within the range between 0 and 1 a decision maker performs the sensitivity analysis of all these weights simultaneously. The difficulty with the Steuer's weight technique is that in many situations a decision maker is unwilling to specify the weights (Lootsma, 1997). Also, the technique is time consuming demanding lot of computation. Lootsma states that apart of wasting of decision maker's time in solving a particular decision problem through the goal programming models, the issue is in a significant degree of decision maker's freedom to select his/her preferences.

In order to eliminate a time consuming component the improvement was suggested to applying the weighted Chebychev norm in a decision process (Benayoun et al, 1971;Yu & Zeleny, 1975; Zeleny, 1974; Kok & Lootsma, 1985). That is to minimise the distance between the objective-function values and so-called ideal values. The technique applied still suffered from the influence of powerful individuals in decision making processes through determination of weights. The computational process was improved through the application of a non-dominated solution where the objective functions were reasonably balanced. That is, "the deviations from the ideal values in the respective directions of optimisation were inversely proportional to the corresponding weights"(Osyczka, 1984). In multi criteria decision analysis the judgmental process is supported by a significant amount of quantitative information, which is called nadir values of the objective functions (Lootsma, 1997). The process of minimisation of nadir vector and ideal vector implies that a decision maker uses of an acceptable compromise under control of weights. In reality, the decision maker cannot always answer the precise questions submitted in the pairwise comparison steps. On the other hand, the nadir-ideal vector is mostly applicable for new decision situations. In the everyday business however, activities are known as being repeatable. With the introduction of the degree of satisfaction with an objective function in the nadir-ideal vector model it is believed that the influence of powerful individuals on decision-making processes was eliminated. For each feasible solution there is a degree of satisfaction with the chosen objective. If the degree of satisfaction is below the nadir value, and above the nadir value, it takes the form of a membership function. By applying a membership function the model takes the fuzzy set approach.

In Osyczka's opinion (1984) the multi criteria optimisation models, being applicable for optimisation of business activities, could be satisfactorily used in a form of linear programming. The business activities' objective functions can be based on weighting coefficients. Managers determine the weighting coefficients on the basis of their intuition, what implies that the weighting coefficients are subject to incomplete information and individual judgement. Weighting coefficients can be presented by a set of weights, which is normalised to sum to 1. Known techniques for comparing this set of weights are eigenvector and weighted least square method (Hwang & Yoon, 1983). The eigenvector technique is based upon a positive pairwise comparison matrix. Since the precise value of two weighting coefficients is hard to estimate, one can use the intensity scale of importance for activities that are broken down into importance ranks. A weighted least square method involves the solution of simultaneous linear equations. Since Bellman and Zadeh's paper in 1970, the maxmin and simple additive weighting method using membership function of the fuzzy set is used in explanation of business decision making problems (Bellman & Zadeh, 1970). Lai and Hwang see the application of fuzzy set theory in decision multi criteria problems as a replacement of oversimplified (crisp) models such as goal programming and ideal nadir vector model.

Fuzzy multi criteria models are robust and flexible. Decision-makers consider the existing alternatives under given constraints, "but also develop new alternatives by considering all possible situations" (Lai & Hwang, 1995).

The transitional step towards fuzzy multi criteria models is models that consider some fuzzy values. Some of these models are linear mathematical formulation of multiple objective decision making presented by mainly crisp and some fuzzy values. Many authors studied such models (Cheng et al, 1999; Chuang et al, 1986; Lai & Hwang, 1993, 1995; Lai, 1995; Zimmerman, 1978). Zimmermann (1978) offered the solution for the formulation by fuzzy linear programming. Lai's (1995) interactive multiple objective system technique contributed to the improvement of flexibility and robustness of multiple objective decision making methodology. Lai considered several characteristic cases, which a business decision maker may encounter in his/her practice. The cases could be defined as both non-fuzzy cases and fuzzy cases. These deal with notions relevant to fuzzy set theory.

## 3. FUZZY LOGIC APPROACH

Fuzzy set theory uses linguistic variables rather than quantitative variables to represent imprecise concepts. Linguistic variables analyse the vagueness of human language.

***Fuzzy set***: Let X be a universe of discourse, $\bar{A}$ is a fuzzy subset of X if for all $x \in X$, there is a number $\mu_{\bar{A}}(x) \in [0,1]$ assigned to represent the membership of $x$ to $\bar{A}$, and $\mu_{\bar{A}}(x)$ is called the membership function of A (Cheng, 1994).

***Fuzzy number***: A fuzzy number $\bar{A}$ is a normal and convex subset of X. Normality implies
$\exists X \in R \lor \mu_{\bar{A}}(x) = 1$.
Convexity implies
$\forall x_1 \in X, \quad x_2 \in X, \quad \forall \alpha \in [0,1]$
$\mu_{\bar{A}}(\alpha x_1 + (1-\alpha) x_2) \geq \min \mu_{\bar{A}}(x_1), \min \mu_{\bar{A}}(x_2)$.

***Fuzzy decision:*** The fuzzy set of alternatives resulting from the intersection of the fuzzy constraints and fuzzy objective functions (Bellman & Zadeh, 1970). A fuzzy decision is defined in an analogy to non-fuzzy environments "as the selection of activities which simultaneously satisfy objective functions and constraints". Fuzzy objective function is characterised by its membership functions. In fuzzy set theory the intersection of sets normally corresponds to the logical "and". The "decision" in a fuzzy environment can therefore be viewed as the intersection of fuzzy constraints and fuzzy objective functions. The relationship between constraints and objective functions in a fuzzy environment is fully symmetric (Zimmerman, 1978).

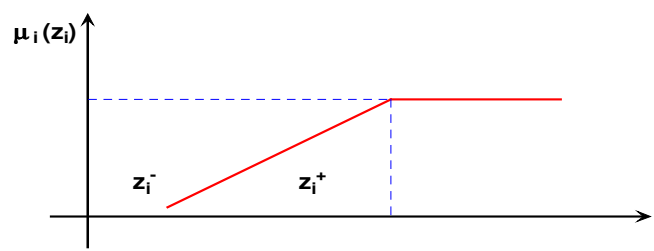

**Figure 1.** Objective function as a fuzzy number

## 4. NON-FUZZY MULTIOBJECTIVE PROBLEM

A general linear multiple criteria decision making model can be presented as:
Find a vector ***x*** written in the transformed form
$$x^T = [x_1, x_2, ...., x_n]$$

which maximises objective functions

$$max \ z_i = \sum_{j=1}^{n} c_{ij} x_j, \quad j=1,2,...,n \qquad (4.1)$$

with constraints

$$\sum_{j=1}^{n} a_{ij} x_j \leq b_i \quad i=1,2,...,m, \ x \geq 0 \qquad (4.2)$$

where $c_{ij}$, $a_{ij}$ and $b_i$ are crisp (non-fuzzy) values. This problem has been studied and solved by many authors. Zimmermann has solved this problem by using the fuzzy linear programming (Zimmermann, 1978). He formulated the fuzzy linear program by separating every objective function $z_i$, its maximum $z_i^+$ and minimum $z_i^-$ value by solving

$$z_i^+ = max\ z_i = \Sigma_j\ c_{ij}x_j, \quad and\quad z_i^- = min\ z_i = \Sigma_j\ c_{ij}x_j, \qquad (4.3)$$

with constraints *(4.2)*. Solutions $z_i^+$ and $z_i^-$ are known as individual best and worst solutions respectively. Since for every objective function $z_i$, its value changes linearly from $z_i^-$ to $z_i^+$ it may be considered as a fuzzy number with the membership function $\mu_i(z_i)$ as shown in Figure 1.

$$\mu_j(z_j) = \begin{cases} 0 & for\ z_i \leq z_i^- \\ \dfrac{(z_i - z_i^-)}{(z_i^+ - z_i^-)} & for\ z_i^- \leq z_i^+, i=1,2,....n \\ 1 & for\ z_i \geq z_i^+ \end{cases} \qquad (4.4)$$

According to Bellman-Zadeh 's principle of decision making in the fuzzy environment the grade of membership of a decision j, specified by objectives $z_i$, is obtained by (Bellmand & Zadeh, 1970)

$$\alpha = min\ \mu_j(z_j), \quad j=1,2,..,k \qquad (4.5)$$

or

$$maxmin\ j$$

subject to

$$\alpha \leq \mu_j(z_j),\ j=1,2,...,k\quad 0 \leq \alpha \leq 1 \qquad (4.6)$$

According to this principle the optimal values of multicriteria optimisation correspond to maximum value of *j*. The auxiliary linear programme is obtained by:

$$z^- = max\ \alpha \qquad (4.7)$$

with constraints *(4.6)*, taking into account *(4.1)* and *(4.4)*

$$-\sum_{j=1}^{n} c_{ij} + (z_j^+ - z_i^-)\alpha \leq \dfrac{(z_i - z_i^-)}{(z_i^+ - z_i^-)} \quad i=1,2,...,k \qquad (4.8)$$

$$0 \leq \alpha \leq 1,\ x_j \geq 0 \qquad j=1,2,..,n$$

The original linear constraints *(4.2)* are added to these constraints. The problem can also be presented in a form (Lai and Hwang 1994): Find a vector *x* subject to

$$z_i(x) \geq\sim z_i^0\ \forall i,\ x \in X \qquad (4.9)$$

where $z_i^0$, $\forall i$ are corresponding goals, and $\geq\sim$ is a soft or quasi inequality. The objective functions are assumed to be maximised

$$max/min\ [\ z_i(x)....z_i(x)\ ] \qquad (4.10)$$

$$x \in X = \{x | g_s(x)\ \{\geq\ =\ \leq\}0,\ s=1,......,m\}$$

where $z_j(x)$, $j \in J$ are maximisation objectives, $z(x)$, $i \in I$ are the minimisation objectives, $I \cup J = \{1,2,..., n\}$ are considered as fuzzy constraints. All functions $z_j(x)$, $g_s(x)$ ($i = 1,...,n$; $s = 1,...,m$) can be linear and nonlinear. With the tolerances of fuzzy constraints given, the membership functions $\mu_i(x)$, $\forall i$ could be established. The feasible set solution obtained through min-operator is defined by interaction of the fuzzy objective set. The feasible set is presented by its membership

$$\mu_D(x) = min(\mu_i(x),..., \mu_k(x),)$$

If a decision maker deals with a maximum $\mu_D(x)$ in the feasible set then the solution procedure is $max(min_i\mu_i(x),)$ $x \in X$. Suppose the overall satisfactory level of compromise is a = $min_i\mu_i(x)$ then the problem can be explained as:

Find max a subject to

$$\alpha \leq \mu_i(x) \; \forall i, \qquad x \in X \qquad (4.11)$$

Assuming that membership functions, based on preference or satisfaction, are linear and non-decreasing between $z_i^+(x)$ and $z_i^-(x)$ for $\forall i$

$$\mu_k(x) = \begin{cases} 1 & if \; z_i(x) \succ z_i^+ \\ \dfrac{(z_i(x) - z_i^-)}{(z_i^+ - z_i^-)} & if \; z_i^-(x) \leq z_i(x) \leq z_i^+ \\ 0 & if \; z_i(x) \prec z_i^- \end{cases} \qquad (4.12)$$

The only feasible solution region is the area $\{x / z_i^-(x) \leq z_i(x) \leq z_i^+\} \forall i$ and $x \in X$. Hence we can write

Find max a subject to

$$\mu_k(x) = [z_i(x) - z_i^-] / [z_i^+ - z_i^-] \geq \alpha \quad x \in X \qquad (4.13)$$

This problem can be solved by using two-phase approach. The first phase relates to the search for an optimal value of $\alpha^0$ in order to find a possible solution ($x^0$). If the possible solution is unique, $x^0$ is an optimal nondominated solution. Otherwise, the second phase is introduced to search for the maximum arithmetic mean value of all membership restricted by original constraints and $\alpha_i \geq \alpha^0 \; \forall$. That is

$$Max \; (\Sigma_i \alpha_i) / i \qquad (4.14)$$
$\alpha' \leq \alpha_i \leq \mu_i(x), \; \forall i, x \in X$

for $i$ objective functions and $\alpha'$ solution *(4.7)*. The objective functions *(4.10)* could be written

$$Max \; [\Sigma_i \mu_i(x)] / I \qquad (4.15)$$
$\alpha' \leq \mu_i(x), \; \forall i, x \in X$

by unifying both objectives *(4.7)* and *(4.11)* the second step can be automatically solved after the first step following the solution procedure of the simplex method:

$$max \; \alpha + \delta[\Sigma_i \mu_i(x)] / I, \qquad \alpha \leq \mu_i(x), \; \forall i, x \in X \qquad (4.16)$$

where $\delta$ is sufficiently small positive number. Since the weights between objectives are not equal we can write

$$max \; \alpha + \delta \Sigma_i w_i \mu_i(x) \quad \alpha \leq \mu_i(x), \; \forall i, x \in X \qquad (4.17)$$

for $w_i$ as the relative importance of the $i^{th}$ objective and $\Sigma_i \mu_i = 1$. The coefficient $\alpha$ represents the degree of acceptability or degree of possibility for the optimal solution. For construction industry activities the minimal value of the coefficient $\alpha_l$ and the maximal value of the coefficient $\alpha_n$ can be prescribed. Hence two new constraints are added in this linear programme:

$$\alpha \geq \alpha_l \quad \alpha \leq \alpha_n \qquad (4.18)$$
where $0 \leq \alpha \leq 1 \quad 0 \leq \alpha,$

Coefficients of satisfaction ($\varphi_i$) in relation to the best individual solutions $z_i^+$ are

$$\varphi_i = max\ z_i / z_i^+ \qquad i=1,2,....,n. \qquad (4.19)$$

Lai and Hwang consider the equation *(4.17)* as an augmented max-min approach that is an extension of Zimmermann's approach. From the aspect of fuzzy set theory the augmented max-min approach allows for compensation among objectives. Firstly one reaches the solutions at a large unit, and then by re-evaluating these solutions the compromise solutions at a smaller unit are obtained.

## 5. FUZZY MULTIOBJECTIVE PROBLEM

The Fuzzy Multiple Objective Decision Model (FMODM) studied by Lai and Hwang (1993, 1995) states that the effectiveness of a decision makers' performance in a decision process can be improved as a result of the high quality of analytic information supplied by a computer. They propose an Interactive Fuzzy Multiple Objective Decision Model (IFMODM) to solve a specific domain of Multiple Objective Decision Model (MODM).

$$Max\ (z_i(x)...., z_n(x)\ ) \qquad (5.1)$$

subject to

$$g_j(x) \leq \sim b_j \quad j=1,...,m \qquad x \geq 0$$

where $b_j$, $\forall j$ are fuzzy resources available with corresponding maximal tolerances $t_i$. Their membership functions are assumed to be non-increasing linear functions between $b_j$ and $b_j + t_j$.

The objective functions *(5.1)* are redefined into
$$Max\ z_i(C_i, x) \qquad i=1,2...,I \qquad (5.2)$$
subject to

$$g_j(A, x)\ \{\leq = \geq\}\ b_j \quad j=1,2,.....m,\ x \geq 0$$

Lai and Hwang (1994) suggest presenting the model *(5.2)* limitations as fuzzy inequalities since the limitations prevent the objective functions from reaching their individual optimum.

Find *x*, subject to

$$z_i(C_i, x) \geq \sim z_i^0,\ \forall i \qquad g_j(Aj,x) \leq b_j,\ \forall_j x \geq 0 \qquad (5.3)$$

where $z_i^0$, $\forall i$ are the goals of the objectives, and $\geq \sim$ is a soft or fuzzy inequality. With the known tolerances of fuzzy constraints the membership functions $\mu_i(z_i)$, $\forall i$ to measure satisfaction levels of fuzzy objective constraints could be established. It is supposed that membership functions are based on a preference concept. The membership functions can be any non-decreasing functions for maximisation objectives and non-increasing functions for maximisation objectives such as linear, exponential, and hyperbolic. Lai and Hwang (1994) assume linear membership functions since the other types of membership functions can be transferred into equivalent linear forms.

Each objective of equation *(5.2)* should have an individual best *($z_i^+$)* and individual worst solution *($z_i^-$)*

$$z_i^+ = max\ z_i(C_i, x),\ x\ \varepsilon\ X$$

$$z_i^- = min\ z_i(C_i, x),\ x\ \varepsilon\ X \qquad (5.4)$$

The linear membership function can be defined as in *(4.8)*. According to *(4.18)* and *(4.19)* the following augmented problem can be defined

$$max\ \alpha + \delta \Sigma_i w_i \mu_i(x) \qquad (5.5)$$

$$\alpha \leq \mu_i(x),\ \forall i,\ x\ \varepsilon\ X,\ \alpha\varepsilon\ [0,1]$$

where $\delta$ is a sufficiently small positive number, and $w_i$ ($\Sigma_i w_i = 1$) is of relative importance or weight. If a decision maker wants to provide his/her goals $z_i^0$ and corresponding tolerances $t_i$ for objectives, than for $z_i^0 \leq z_i^+$ and $(z_i^0 - t_k)b \geq z_i^-$ the problem will become:

Find *x*, subject to

$$z_i (C_i,x) ) \gtrsim z_i^0, \forall i \quad \text{and } x \varepsilon X \quad (5.6)$$

where $z_i^0$, $\forall i$ as well as their tolerances $t_i$ are given. Then

$$max\ \alpha + \delta \Sigma_i w_i \mu_i (z_i)$$
$$\mu(z_i) = 1 - [z_i^0 - z_i (C_i,x)]/ t_i \geq \alpha \quad x \varepsilon X, \alpha \varepsilon [0,1] \quad (5.7)$$

The problem can be further considered as

$$max\ \alpha + \delta[\Sigma_i w_i \mu_i (z_i) + \delta \Sigma_j q_j \mu_j (g_j) ] \quad (5.8)$$
subject to

$$\mu_i (z_i) = [z_i (C_i,x) - z_i^- ]/ [z_i^+ - z_i^-] \geq \alpha\ \forall i$$

$$\mu_j (g_j) = 1- [g_j(A_j,x)- b_j ]/ t_j \geq \alpha\ \forall j \quad x>0, \alpha \varepsilon [0,1]$$

where $w_i$ and $g_j$, $\forall i, j$ are of relative importance and $\Sigma_i w_i + \Sigma_j g_j = 1$

According to this procedure the computer programme has been written in FORTRAN 77 programming language. Input data are: number of objectives *k*, number of constraints *m*, number of unknowns *n*, goals $z_i$ (i=1,2,..,k), elements $c_{ij}$ (i=1,2,..,k; j=1,2,...,n), $a_{ij}$ (i=1,2,..,n), $b_i$ (i=1,2,..,m), tolerances $t_i$ (i=1,2...,k) and $d_i$ (i=1,2,...,m). The programme determines the individual best $z_i^+$ solution and the individual worst solution $z_i^-$ for every objective *i* (i=1,2,...,k). The objective functions are *(4.3)* and the constraints are *(4.2)*. The obtained values $z_j^+$ and $z_j^-$, based on the modified Zimmermann's procedure, are used to solve the linear programme with the objective function *(4.17)* and constraints *(4.2)*, *(4.8)* and *(4.18)*. For the nonfuzzy problem, this programme gives the values of unknown $x_j$ (j=1,2,..,n), maximal values of objective function $z_i$ (i=1,2,...,k), coefficient of acceptability $\alpha$ and coefficients of satisfaction $\varphi_i$ (i=1,2,..,k). For the fuzzy problem, the linear programme with the objective function *(5.3)* and the constraints *(5.6)* gives: the optimal value of unknown $x_i$ (i=1,2,...,n), objective function $z_i$, coefficients of satisfaction $\varphi_i$ (i=1,2,..,k) and coefficient of acceptability $\alpha$.

## 6. CASE STUDY ANALYSIS AND MODELLING

The operations of a concrete manufacturing plant, which produces and transports concrete to building sites, have been analysed. Fresh concrete is produced at a central concrete plant and transported by seven transit mixers over the distance ranging *1500-3000 m* (depending on the location of the construction site) to the three construction sites. Three concrete pumps and eleven interior vibrators are used for delivering, placing and consolidating the concrete at each construction site. Table 1 illustrates the manufacturing capacities of the plant, operational capacity of the concrete mixer, interior vibrator, pumps and manpower requirement at the three construction sites. A quick analysis will reveal the complexity of the variables and constraints of this concrete production plant and delivery system. The plant manager's task will be to optimise the profit by utilizing the maximum plant capacity while meeting the three-construction site's concrete and other resource requirement through a feasible schedule.

**Table 1:** Concrete plant capacity and construction site's resource demands

|  | Concrete Plant | Site A | Site B | Site C | Remarks |
|---|---|---|---|---|---|
| Plant capacity | 60 m³/h 2520m³ (weekly) | | - | | 200 m³ (tolerance) |
| Transit mixers (total = 7) | | 8.45 m³/h | 9.26 m³/h | 7.26 m³/h | operated by 7 workers |
| Concrete pumps (total = 3) | - | 16 m³/h | 22 m³/h | 26 m³/h | operated by 6 workers |
| Interior vibrators (total = 11) | | 4.0 m³/h | | | |
| Worker requirement | 5 | 6 | 7 | 9 | |
| Minimal concrete requirement (tolerance) | | 14.0 m³/h 588 m³/week (47 m³) | 18.0 m³/h 756 m³/week (60 m³) | 21.5 m³/h 903 m³/week (72m³) | |
| Weekly values are based on 42 working hours/week | | | | | |

### 6.1. Objective Formulation

Success of any decision model will directly depend on the formulation of the objective function taking into account all the influential factors. We modelled the final objective function taking into account three independent factors: (1) profit expressed as $/m³ (2) index of work quality (performance) and (3) worker satisfaction.

***Profit:*** The expected profit as related to the volume of concrete to be manufactured is modelled as the *first objective* and is shown in Table 2. The minimal expected weekly profit as a fuzzy value is $z^0 = AU\$27,000$ per week with tolerance, $t_1 = AU\$2,100$.

**Table 2:** Modelling profit as an objective

|  | Site A | Site B | Site C |
|---|---|---|---|
| **Expected profit (AU$/m³)** | 12 | 10 | 11 |

***Index of quality:*** Equally or sometimes more important than the profit, quality plays an important role in every industry. We modelled the index of quality at construction sites, as the *second objective*. The index is ranged from *5* points/m³ (bad) quality to *10* points/m³ (excellent) quality and the assigned values are shown in Table 3. The minimal expected total weekly number of points for quality, as fuzzy value, is $z^0_2 = 21400$ with tolerance, $t_2 = 1700$ points.

**Table 3:** Modelling index of quality as an objective

|  | Site A | Site B | Site C |
|---|---|---|---|
| **Index of Quality** | 9 | 10 | 7.5 |

***Worker Satisfaction Index:*** We modelled the index of worker satisfaction as the *third objective* and is ranged from *5* to *10* points per $m^3$ of produced, transported and placed concrete. The assigned values are depicted in Table 4. The minimal expected total weekly number of points as a fuzzy value is $z^0 = 18000$ with tolerance, $t_3 = 1400$.

**Table 4:** Modelling worker satisfaction index as an objective

|  | Site A | Site B | Site C |
|---|---|---|---|
| **Worker Satisfaction Index** | 9 | 10 | 7.5 |

## 6.2. Variables that Optimise the Objective Function

After knowing the objective function our next task is to determine the variables that optimises the objective function. In our problem it is to find: the optimal value of unknowns $x_i$ $(i=1,2,3)$ that represent quantities of concrete which have to be delivered to Site A, B and C respectively and corresponding optimal values of the objective functions $z_1, z_2, z_3$. According to problem requirements and available data (Table 1,2,3 and 4) the objective functions can be modelled as follows (Zimmermann, 1978):

- *max $z_1=12x_1+10x_2+11x_3$ ($>,\sim$) 27000   with tolerance, $t_1=2100$ (profit)*
- *max $z_2=9x_1+10x_2+ 7.5x_3$ ($>,\sim$) 21400 with tolerance, $t_2=1700$ (index of quality)*
- *max $z_3= 8x_1+7x_2+9x_3$ ($>,\sim$) 18000 with tolerance, $t_3=1400$.(worker satisfaction index)*
- *$x_1+x_2+x_3$ ($<,\sim$) 2520, tolerance $d_1=200$ (weekly capacity of the concrete plant)*
- *0.12 $x_1$ + 0.11 $x_2$ + 0.14 $x_3$ ($\leq,\sim$) 7x42= 294 h, tolerance $d_2=23$ h (weekly engagement of 7 transit mixers, taking into account of their working capacities)*
- *0.06 $x_1$ +0.05 $x_2$ + 0.04 $x_3$ ($<,\sim$) 3x42=126h, tolerance $d_3=10h$ (weekly engagement of 3 concrete pumps)*
- *$6x_1+7x_2+9x_3$($<,\sim$) 22x42=924, tolerance $d_4=74$. (weekly engagement of 22 workers for interior delivering, placing and consolidating concrete at sites A,B and C).*
- Minimal weekly requests for concrete from the three construction sites:
  - Site A, $x_1 \geq 588\ m^3$, tolerance $d_5=47m^3$
  - Site B, $x_2 \geq 756\ m^3$, tolerance $d_6=60\ m^3$
  - Site C, $x_3 \geq 756\ m^3$, tolerance $d_7=72\ m^3$
- The minimal value of the degree of acceptability is $\alpha_1 \geq 0.80$. These constraints written in full are:
  - *$x_1+x_2+x_3$($<,\sim$)2520*
  - *$0.118x_1+0.108x_2+0.139x_3$ ($<,\sim$) 294*
  - *$0.063x_1+0.045x_2+0.038x_3$ ($<,\sim$) 126*
  - *$0.100x_1+0.117x_2+0.150x_3$ ($<,\sim$) 924*
  - *$0.033x_1+0.033x_2+0.055x_3$ ($<,\sim$ ) 294*
  - *$x_1$($>,\sim$)588     $x_2$($>,\sim$)756 $x_3$ ($>,\sim$)903.*

By using linear programming technique we will be able to solve the above equations and the individual best and worst non-fuzzy solution for constraints (b) and individual objective functions (a) could be obtained. The obtained results are summarized in Table 5.

## 6.3. Solutions

Now we will try to solve the multiple objective functions using the results obtained for $z_i^+$ and $z_i^-$ as shown in Table 5 and using the modified Zimmermann's procedure as discussed in Section 4. We implemented the codes in FORTRAN 77 and were executed in a Windows NT, Pentium II Machine. The results obtained are summarized in Table 6. The simulations were repeated three times and found that the results are stable. Coefficient of acceptability of this solution was found to be $\alpha=0.941$. When the objective functions were modelled using the described fuzzy approach the obtained solutions are as summarized in Table 7.Coefficient of acceptability of this solution $\alpha =0.852$.As depicted in Figures 2 and 3, the obtained results clearly shows the superiority of fuzzy approach. However it is also interesting to note that there is not much difference between fuzzy and non-fuzzy solutions for the three objective functions. The difference is being less that *2* percent. The coefficients of acceptability of the solutions $\alpha$, indicating the possibility of realising these solutions, are very high. According to this, the decision maker could accept

- the non-fuzzy solution that gives smaller profit with possibility of realisation $\alpha=0.941$

- the fuzzy solution that gives higher profit with possibility of realisation $\alpha=0.852$

Table 5. Individual best and worst non-fuzzy solution

| Objective | $x_1$ (m³/week) | $x_2$ (m³/week) | $x_3$ (m³/week) | $z_i^+$ ($) | $z_i^-$ ($) |
|---|---|---|---|---|---|
| 1 | 734.02 | 756.00 | 903.00 | 26301.29 | 0 |
| 2 | 588.00 | 915.95 | 903.00 | 21224.00 | 0 |
| 3 | 734.02 | 756.00 | 903.00 | 19291.00 | 0 |

Table 6. Optimal results using non-fuzzy procedure

| $x_1$ (m³/week) | $x_2$ (m³/week) | $x_3$ (m³/week) | Max ($z_1$) $\varphi^*$ | Max ($z_2$) $\varphi^*$ | Max ($z_3$) $\varphi^*$ |
|---|---|---|---|---|---|
| 635.94 | 863.43 | 903.0 | 26,199 | 21,130 | 19,259 |
|  |  |  | 0.996 | 0.996 | 0.998 |

- $\varphi$ is the coefficient of satisfaction

(a)

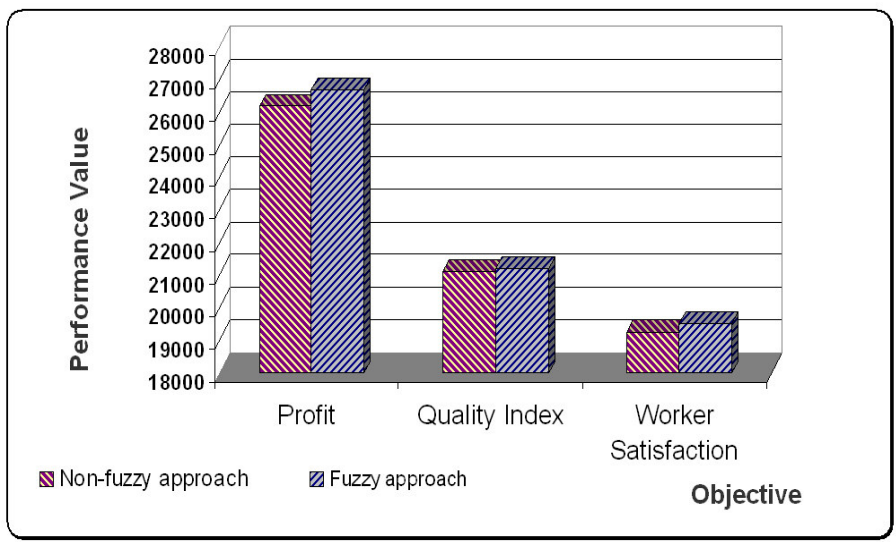

(b)

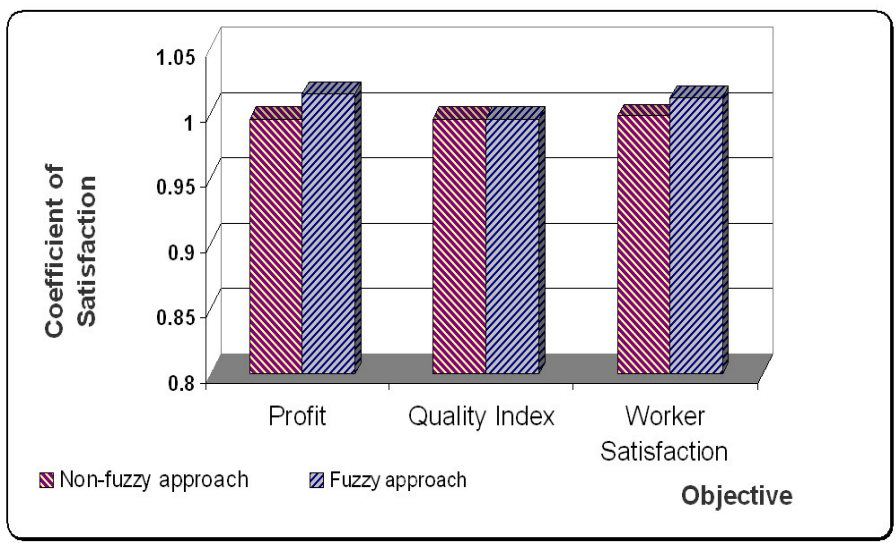

**Figure 2.** Comparison of fuzzy and non-fuzzy approach (a) showing the performance value of objective functions (b) showing the coefficient of satisfaction for the different objective functions.

The developed software will also help the decision maker to vary the values of the coefficient $\alpha$ in the interval $[0,1]$ and to receive the corresponding optimal values of production and profit with corresponding values of possibility. After careful study of the optimal values of the objective functions and the various constraints, very often expert problem domain knowledge will be required to understand the possibility of realization of the achieved results.

## 7. CONCLUSIONS

Decision making processes that support organising business activities, being subject to imprecise data and subjective judgements, can be explained by the fuzzy approach of the multiple objective criteria models. The analysis and modelling of the construction industry problem presented in this paper is based on both linear objective functions and constraints in a form of linear membership functions. The results clearly indicate the superiority of the fuzzy approach in terms of best individual solutions for the three objective functions and coefficient of satisfaction. The difference between fuzzy and non-fuzzy objective functions for the individual best solutions is less than 2%. However, the possibility of realisation of the proposed fuzzy approach was 10% less than the non-fuzzy approach. Hence there seems to be a trade off between the possibility of realization and optimal profit. The developed software is also capable of computing the optimal profit for a given possibility of realisation coefficient.

The work also implies that the membership functions once introduced in a linear programming model either in constraints, or both in objective functions and constraints, provide a similar (or even better) level of satisfaction for obtained results. Like in any optimisation problem, success directly depends on modelling the objective functions of the problem concerned and the modelling of the variables. For the future research it is suggested that the proposed models could be improved with the constraint programming in order to find out whether such models could support decisions making processes in industry' activities other than construction. Using some adaptive computation techniques and the proposed fuzzy technique, it will be also interesting to find out the optimal values for the objective functions for the possibility of realization coefficient similar or even more than the non-fuzzy approach.